# A Lane-Changing Prediction Method Based on Temporal Convolution Network

Yue Zhang
Key Laboratory of Road and Traffic Engineering of Ministry of Education
Tongji University, Shanghai 201804, China
E-Mail Address: zhangyue18@tongji.edu.com

Yajie Zou
(corresponding author)
Key Laboratory of Road and Traffic Engineering of Ministry of Education
Tongji University, Shanghai 201804, China
E-Mail Address: yajiezou@hotmail.com

Jinjun Tang
School of Traffic & Transportation Engineering
Central South University, Changsha, 410075, China
E-Mail Address: jinjuntang@cus.edu.cn

Jian Liang
School of Traffic & Transportation Engineering
Central South University, Changsha, 410075, China
E-Mail Address: m18373162995_1@163.com

**Abstract**

Lane-changing is an important driving behavior and unreasonable lane changes can result in potentially dangerous traffic collisions. Advanced Driver Assistance System (ADAS) can assist drivers to change lanes safely and efficiently. To capture the stochastic time series of lane-changing behavior, this study proposes a temporal convolutional network (TCN) to predict the long-term lane-changing trajectory and behavior. In addition, the convolutional neural network (CNN) and recurrent neural network (RNN) methods are considered as the benchmark models to demonstrate the learning ability of the TCN. The lane-changing dataset was collected by the driving simulator. The prediction performance of TCN is demonstrated from three aspects: different input variables, different input dimensions and different driving scenarios. Prediction results show that the TCN can accurately predict the long-term lane-changing trajectory and driving behavior with shorter computational time compared with two benchmark models. The TCN can provide accurate lane-changing prediction, which is one key information for the development of accurate ADAS.

**1. Introduction**

Autonomous vehicles can effectively ease traffic congestion, improve traffic safety and reduce vehicle emission pollution. One challenging task of autonomous driving is to model the driving behavior and then predict the future vehicle movement. As one of the basic driving behaviors, lane change is a complex vehicle movement and can potentially result in serious traffic accidents. Lane change includes vehicle movement in both horizontal and vertical directions, and frequent lane change behaviors can significantly affect traffic flow operation. The National Highway Traffic Safety Administration (NHTSA) estimates that in 2007, among all traffic accidents reported by the police, lane change and merging traffic accidents accounted for about 0.5% of traffic fatality (Guo, et al., 2010). In addition, studies have shown that dangerous lane changes can cause unstable traffic flow (Yang, et al., 2009), and the lane change/merging accidents can also cause significant traffic delay (Chovan, et al., 1994). Modeling the lane-changing process is useful for Advanced Driver Assistance System (ADAS) (Nilsson, et al., 2017) to implement automatic emergency braking (AEB), forward collision warning (FCW) and lane departure warning (LDW), etc. Therefore, accurate prediction models are definitely useful to improve the safety of lane-changing behavior.

The lane-changing process can be divided into lane-changing behavior and lane-changing trajectory. Lane-changing behavior is driver's immediate operation of the vehicle, such as steering wheel angle and braking force (Salvucci, et al., 2007; Schmidt, et al., 2014b; Tang, et al., 2018; Tang, et al., 2019). Lane-changing trajectory can reflect the driving behavior during the lane-changing process, such as the position and acceleration of vehicles (Chen, et al., 2019; Liu, et al., 2014; Nishiwaki, et al., 2009; Tomar & Verma, 2011; Xu, et al., 2012; Yao, et al., 2013; Zhou, et al., 2017). The current lane-changing prediction studies can be generally divided into two categories: lane-changing behavior prediction and lane-changing trajectory prediction.

**1.1 Lane-changing behavior prediction**

The traditional method is to use turn signal as the only indicator to reflect lane-changing intention (Ponziani, 2012), while this method lacks sensitivity and specificity to predict lane-changing behavior (Schmidt, et al., 2014a). Nowadays, advanced automatic detection technology can collect multimodal data for driving behavior analysis, such as electroencephalogram, eye-tracking, steering angle, braking force, obstacle distance, etc. Based on these data, some lane-changing prediction techniques were developed, which include Hidden Markov model (HMM), fuzzy logic system, regression model, and artificial neural network. HMM is a time-series probabilistic model, which can capture the hidden state of the input sequence to predict driving behaviors. Yuan, et al. (2018) used the HMM to predict lane-changing maneuver of the vehicle ahead for adaptive cruise control system.

Fuzzy logic system can predict the driving behavior during lane changes by establishing the probability reasoning process of fuzzy input parameters and optimizing the parameters in the fuzzy membership function. Shi and Zhang (2013) proposed fuzzy logic method to analyze multi lane-changing behavior and adopted steering wheel angle as the output variable to evaluate the efficiency of lane-changing process. In addition, the regression model was often used to explore the influence mechanism of driving behavior. Henning, et al. (2007) used regression model to estimate the intention of lane changes considering some influencing factors such as glance to the left outside mirror, turn signal, and lane crossing. In order to further improve the prediction accuracy, many researchers have introduced the artificial neural network with the strong learning ability and transfer ability to predict driving behavior. Díaz, et al. (2018) predicted lane-changing action considering speed, acceleration and jerk with the multilayer perceptron (MLP) and convolutional neural networks (CNN) architectures.

**1.2 Lane-changing trajectory prediction**

There are two common approaches to address the lane-changing trajectory modeling, i.e., kinematic models and stochastic models. Kinematic models which describe the lane-changing maneuver in a form of mathematical equations (e.g., lateral acceleration versus time) can be listed as follows: polynomials (Yao, et al., 2012, 2013), sinusoidal (Butakov

& Ioannou, 2015) and trapezoidal (Kanaris, et al., 2001) functions, etc. In general, these models can provide smoothing lane-changing trajectories and well describe the shape of lane-changing process. However, most of them cannot be used to interpret drivers' lane-changing behavior well (Xie, et al., 2019) or consider the heterogeneity of the lane changes and interaction with the surrounding environment. On the one hand, these models provide meaningful kinematic information about the maneuver, they can describe lane-changing trajectories in the long term. On the other hand, their rigid structure does not take into account all aspects of driving behavior, such as dependence of trajectory shape on speed and surrounding traffic configuration. In addition, a system that depends on a human response is nondeterministic by nature and driver's uncertainty could cause longer reaction times (Petzoldt & Krems, 2014).

On the contrary, stochastic models usually extract uncertain lane-changing characteristics from massive data. Moreover, the stochastic models can directly describe human driving behavior without any restriction imposed by the kinematic models. Common modeling techniques include Stochastic Switched AutoRegressive eXogenous model (SS-ARX) (Celik & Ertugrul, 2010), HMM (Liu, et al., 2014), neural network, fuzzy system (Hou, et al., 2012), Bayesian network (Malta, et al., 2008), support vector machine (Dou, et al., 2016), and Gaussian mixture model (Wiest, et al., 2012). Among them, neural network is the most widely used model due to its flexible structure and accurate prediction performance. Ding, et al. (2013) developed a two-layer Tansig and Linear Backpropagation (BP) neural network model to predict lane-changing trajectories in a real-time manner. In Yoon and Kum (2016), a lane-based probabilistic lateral motion prediction algorithm was proposed based on MLP approach. Based on the historical data, the proposed MLP model can estimate the future vehicle lane change trajectory. The lane changing process shows time-series characteristics, and the future vehicle movement is closely related to the past states. The recurrent neural network (RNN) (Chen, 2018; Xie, et al., 2019) model takes advantage of considering historical sequence but is still difficult to scale to very long data sequences. But these models generally provide unsatisfactory long-term lane-changing prediction results. However, a long-term lane-changing trajectory prediction can provide sufficient time for drivers to make reasonable driving decisions. In order to obtain long-term vehicle trajectory, some researchers try to use the hybrid model (Butakov & Ioannou, 2015; Xie, et al., 2017) to combine the advantages of kinematic and stochastic methods.

Although different studies have been proposed to predict lane-changing behavior and trajectory, there still exist some gaps need to be further explored. The lane-changing behavior is a complicated process. The existing lane-changing process prediction models often require a significant amount of computational time and the long term prediction results are usually unsatisfactory (Bai, et al., 2018), which is not applicable for real time driving assistance. To address above problems, this paper proposes a temporal convolutional network (TCN) to predict the lane-changing behavior and trajectory. Specifically, an advanced deep learning approach (i.e., the TCN) is adopted to model the lane-changing process. Compared with the RNN and CNN, the TCN considers a significant amount of historical data, and provides a more accurate long-term lane-changing process prediction with less computational time. This paper proposes the TCN which is a stochastic approach with the consideration of the time-series characteristics of lane-changing maneuver. The proposed model can accurately predict the long-term lane-changing behaviors and trajectories. Moreover, the model can be easily trained using driving simulation data and then be applied in ADAS with quick responsiveness.

## 2. Methodology

Since the lane-changing process shows time-series characteristics, it is necessary to consider the historical driving data when modeling the current and future lane-changing behaviors or trajectories. With the development of vehicle detection system, a significant amount of driving data can be collected during a lane-changing event. To capture the long-range amount of historical data, the TCN with a hierarchy of temporal convolutional filter is introduced.

### 2.1 The TCN structure

Compared with other convolutional architectures for the sequential data, the TCN is designed from the first principle to combine simplicity, autoregressive prediction, and very long memory. The structure is described as follows.

#### 2.1.1 Sequence Modeling and Causal Convolutions
The nature of sequence modeling task is that the output sequence only depends on the input sequence at the current and previous times instead of any future inputs as shown in Figure 1. Formally, a sequence modeling network is any function $f: X^{T+1} \rightarrow Y^{T+1}$ that produces the mapping

$$\hat{y}_0, \cdots, \hat{y}_T = f(x_0, \cdots, x_T) \tag{1}$$

Where $x_0, \cdots, x_T$ are the given input sequence and $\hat{y}_0, \cdots, \hat{y}_T$ are the corresponding outputs at each time. To achieve no leakage from the future into the past, the TCN uses causal convolutions, convolutions where an output at time *t* is convolved only with elements from time *t* and earlier in the previous layer. A major disadvantage of this basic design is that in order to achieve a long effective history size, we need an extremely deep network or very large filters.

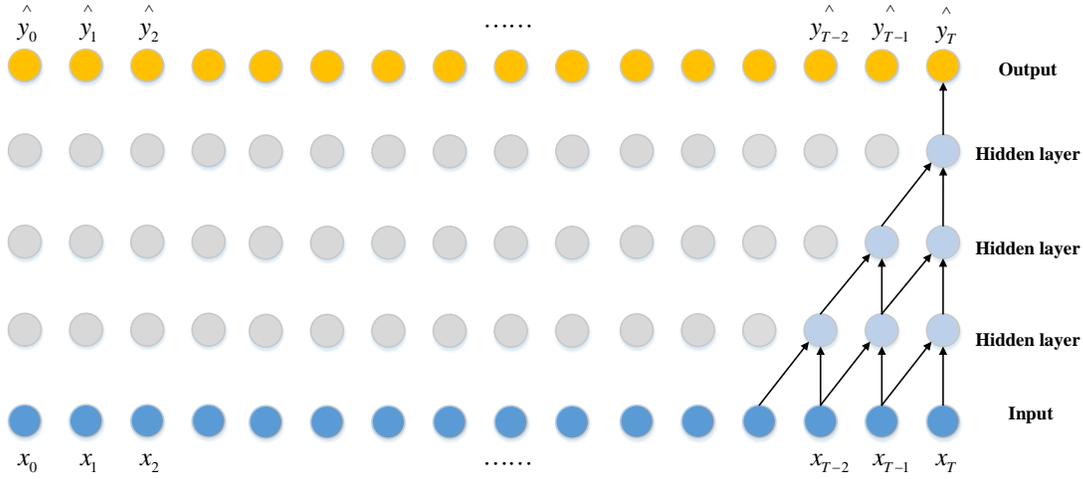
Figure 1   Visualization of a stack of causal convolutional layers

### 2.1.2 Dilated Convolutions

A simple causal convolution is only able to look back at a history with size linear in the depth of the network. This makes it challenging to apply the aforementioned causal convolution on sequence tasks, especially those requiring longer history. The solution is to employ dilated convolutions that enable an exponentially large receptive field. Figure 2 displays the structure of dilated convolution layers. For a 1-D sequence input $x \in R^n$ and a filter $f:\{0,\cdots,k-1\} \to R$, the dilated convolution operation $F$ on element s of the sequence is defined as Equation 2.

$$F(s) = \sum_{i=0}^{k-1} f(i) \cdot x_{s-d \cdot i} \tag{2}$$

Where $d$ is the dilation factor, $k$ is the filter size, and $s - d \cdot i$ represents the direction of the past.

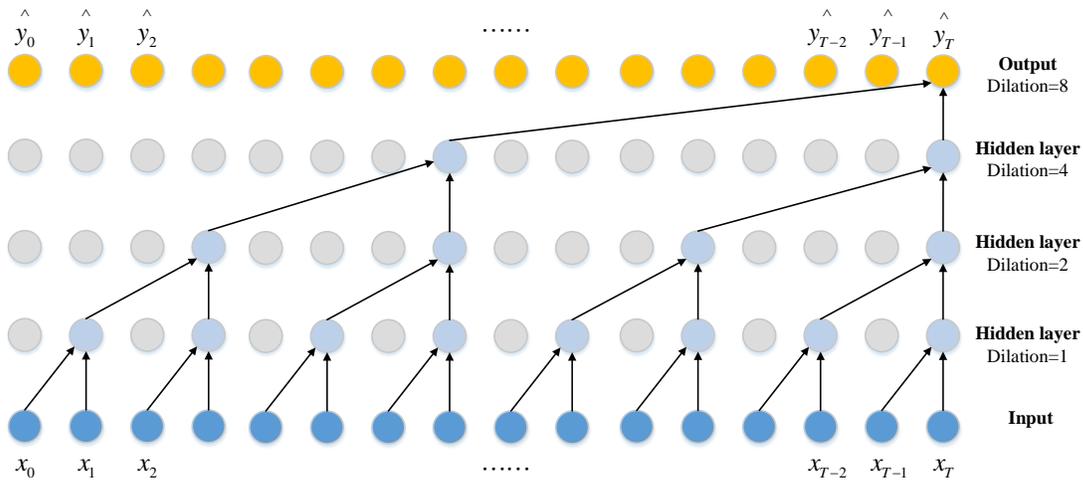
Figure 2   Visualization of a stack of dilated convolutional layers (k=2, dilations=[1,2,4,8])

### 2.1.3 Residual Connections

The TCN employs a generic residual module (He, et al., 2016) instead of a convolutional layer. The residual block for TCN is shown in Figure 3. The residual structure replaces simple connection between layers in the TCN (and ConvNets in general) and increases stabilization of deeper and larger TCNs. To account for discrepant input-output widths in the TCN, an additional 1x1 convolution is used to ensure that element-wise addition $\oplus$ receives tensors of the same shape.

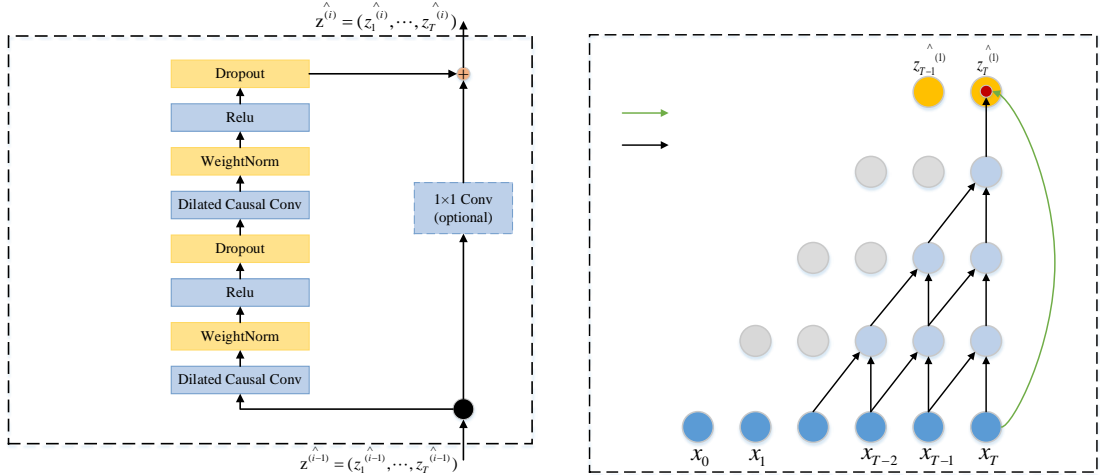

Figure 3 a) The TCN residual block. b) An example of residual connection in a TCN.

Compared to other artificial neural networks, the distinguishing characteristics of TCN are: 1) the convolutions in the architecture are causal, meaning that there is no information "leakage" from future to past; 2) the TCN could build very long effective historical dependence (i.e., the TCN can consider extensive long historical data as the inputs) using a combination of deep networks (augmented with residual layers) and dilated convolutions.

### 2.2 Lane-changing prediction model

In order to describe lane-changing prediction model using the TCN structure, we develop the TCN-based lane-changing prediction model. According to the literature review, the steering wheel angle $\alpha$ is selected to represent the lane-changing behavior (Tang, et al., 2018; Tang, et al., 2019), and the two-dimensional positions (longitudinal position $x$ and lateral position $y$) of the vehicle represent the lane-changing trajectory, see Figure 4. The input of the joint model comprehensively consider the impact of historical lane-changing behavior and trajectory.

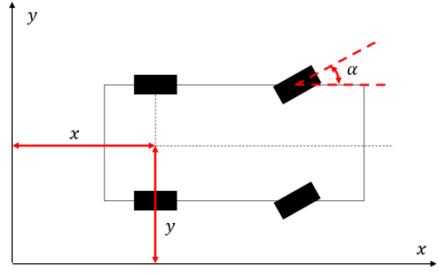

Figure 4  A schematic diagram of vehicle parameters

## 2.3 Input and output

Let $X_t \in \mathbb{R}^D$ be the D-dimension vector of the *t*-th time step and $X_t$ is composed of the position and steering wheel angle of subject vehicle. The composition of the feature vector $X_t$ in *t*-th time step is $X_t = \{\alpha_t, x_t, y_t\}$. Thus, the input matrix $X_t$ of model training and prediction is composed of feature vectors in L time steps $\Delta t$ before *t*-th time step. The expression for $X_t$ is as follows:

$$X_t = \{X_{t-L\Delta t}, X_{t-(L-1)\Delta t}, \cdots, X_{t-2\Delta t}, X_{t-\Delta t}\} \tag{3}$$

The existing studies on lane-changing prediction consider many factors such as speed, acceleration, time gap, space gap, etc. However, the feature vector $X_t$ in this study is composed of information of steering wheel angle and position at multiple moments, which means the TCN can be used to characterize these parameters during continuously detecting. This kind of inputs ensure the maximum flexibility for the vehicle's detecting capabilities if consider how to implement for autonomous vehicles in future (Zhang, et al., 2019) .

The model output $\hat{Y}_t$ corresponding to the input $X_t$ at *t*-th time step is the predictive position and steering wheel angle of the vehicle in *T* prediction horizon, which is described as Equation 4.

$$\hat{Y}_t = \{\hat{X}_t, \hat{X}_{t+\Delta t}, \cdots, \hat{X}_{t+(T-1)\Delta t}, \hat{X}_{t+T\Delta t}\} \tag{4}$$

Finally, each sample of the input sequences $X_t$ is standardized using Min-Max normalization which brings $X_t$ to within a standard [-1, 1] range.

$$X_t^* = \frac{X_t - \min(X_t)}{\max(X_t) - \min(X_t)} \tag{5}$$

where $X_t^*$ is the final input for the TCN.

### 2.4 Configuration of the TCN

This section describes the optimization process for determining the optimal configuration settings of the TCN.

#### 2.4.1 Loss function

The training process of TCN is essentially a back propagation process of the loss function, which is used to adjust the weights and parameters. Through training, the loss function is gradually reduced and the output is gradually close to the true value until the optimal model is obtained. Common loss function used in the neural network include mean squared error (MSE), cross entropy, KL divergence, hinge loss, etc. To evaluate the model performance, MSE which is the most commonly used loss function in regression is adopted. For sequence modeling tasks, the expression of MSE is defined in Equation 6.

$$MSE = \frac{\sum_{k=1}^{N} \sum_{t=1}^{T} \left[ \left( \hat{Y}_{ti} - Y_{ti} \right)^2 \right]}{T \times N} \tag{6}$$

where $N$ is the total sample size, $T$ is prediction horizon, $\hat{Y}_{ti}$ is the $i$-th sample calculated by the model, and $Y_{ti}$ is the observed value at $t$ time step.

#### 2.4.2 Activation function

Artificial neural network introduces activation function to improve the nonlinear modeling ability so that the network could solve more complex problems. Common activation function used in artificial neural network are sigmoid function, tanh function, rectified linear unit (ReLU) function, etc. Since the ReLU function has the characteristics of fast convergence, ease of gradient disappearance, and simple function form, this study adopts the ReLU function in the TCN. The mathematical expression of ReLU is shown in Equation 7.

$$ReLU(x) = \begin{cases} x & if\ x > 0 \\ 0 & if\ x \leq 0 \end{cases} \tag{7}$$

#### 2.4.3 Dropout

Dropout means that during the training process of the deep learning network, some neural network units are temporarily dropped from the network according to a certain probability. This is a common method used to prevent the model from overfitting. The value of dropout rate is set as 0.1 according to the results of parameter adjustment.

#### 2.4.4 Optimization algorithm

Optimization algorithm determines the way to iteratively optimize the loss function of the network. A method of Adaptive Moment Estimation (Adam) optimizer is used and the steps for weight update are shown in Equations 8-13. First, the gradient of loss function $f_t(\theta)$ is determined as Equation 8.

$$g_t = \nabla_\theta f_t(\theta) \tag{8}$$

Then the first moment and second raw moment are calculated as Equations 9 and 10, which is similar to a moving average so that each update is related to the historical value.

$$m_t = \beta_1 m_{t-1} + (1 - \beta_1)g_t \qquad (9)$$
$$v_t = \beta_2 v_{t-1} + (1 - \beta_2)g_t^2 \qquad (10)$$

where $m_t$ is the first moment of the gradient at t-th step and $v_t$ is the second raw moment at $t$-th step. $\beta_1$, $\beta_2$ are exponential decay rates for the moment estimates and $\beta_1, \beta_2 \in [0,1)$. Next, compute bias-corrected first moment estimate and bias-corrected second raw moment estimate, which are shown in Equations 11 and 12.

$$\hat{m}_t = \frac{m_t}{1 - \beta_1^t} \qquad (11)$$

$$\hat{v}_t = \frac{v_t}{1 - \beta_2^t} \qquad (12)$$

As the value of $t$ increases, $1 - \beta_1^t \to 1$ and $1 - \beta_2^t \to 1$, which indicates the completion of the bias correction task. Finally, update parameter $\theta_t$ as Equation 13.

$$\theta_{t+1} = \theta_t - \frac{\eta}{\sqrt{\hat{v}_t} + \varepsilon} \hat{m}_t \qquad (13)$$

where $\eta$ is learning rate used to control the step size and is set as 0.001 in this study. Besides, other constant parameters are set as default values, such as $\beta_1 = 0.9$, $\beta_2 = 0.999$ and $\varepsilon = 10^{-8}$.

## 3. Data Description

The data used in this study are collected by a driving simulator shown in Figure 5. Note that this driving simulation data have been used in the previous studies (Tang, et al., 2018; Tang, et al., 2019). The simulated scenario is that a lead vehicle is traveling in the front of subject vehicle with constant speed a two-lane road for one direction. And the lead vehicle is a van (6915 mm in length, 2150 mm in width and 2260 mm in height) and lane width is 3.5m. A total of 47 experienced drivers (22 females and 25 males) ranging from 29 to 47 years old need to successfully complete three valid lane-changing tasks in scenarios where the lead vehicle runs at speed levels of 60 km/h, 80 km/h and 100 km/h respectively. The data are recorded at a frequency of 60hz (i.e., a time step is 0.017 second).

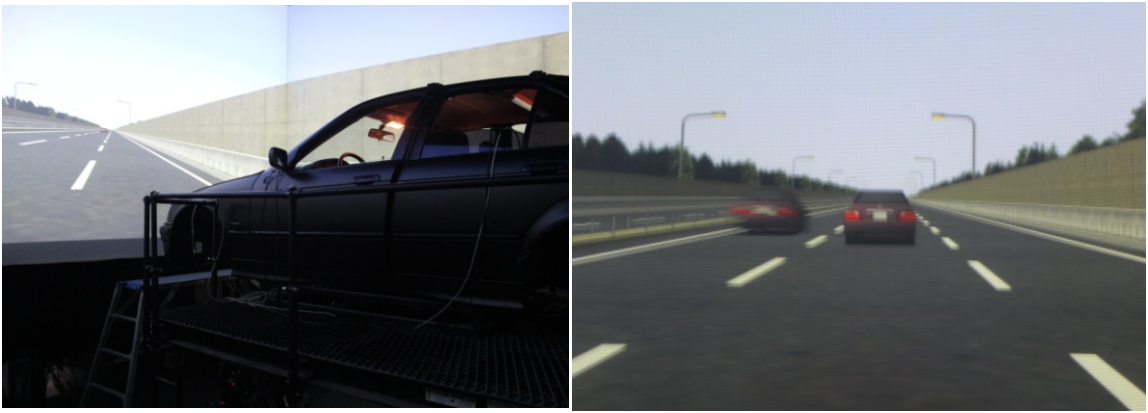

a simulation vehicle                          b simulation scenario

Figure 5   Driving simulator used in the study

A data sample for one lane-changing event is demonstrated in Figure 6. As shown in Figure 6, the subgraph in the first row shows a lane-changing trajectory in a relative coordinate system and the residual subgraph shows the distribution of $\Delta v$, $\Delta x$, $\Delta y$ and $\alpha$ respectively. Note that compared with other variables, the lane-changing behavior variable (i.e., steering wheel angle $\alpha$) fluctuates significantly.

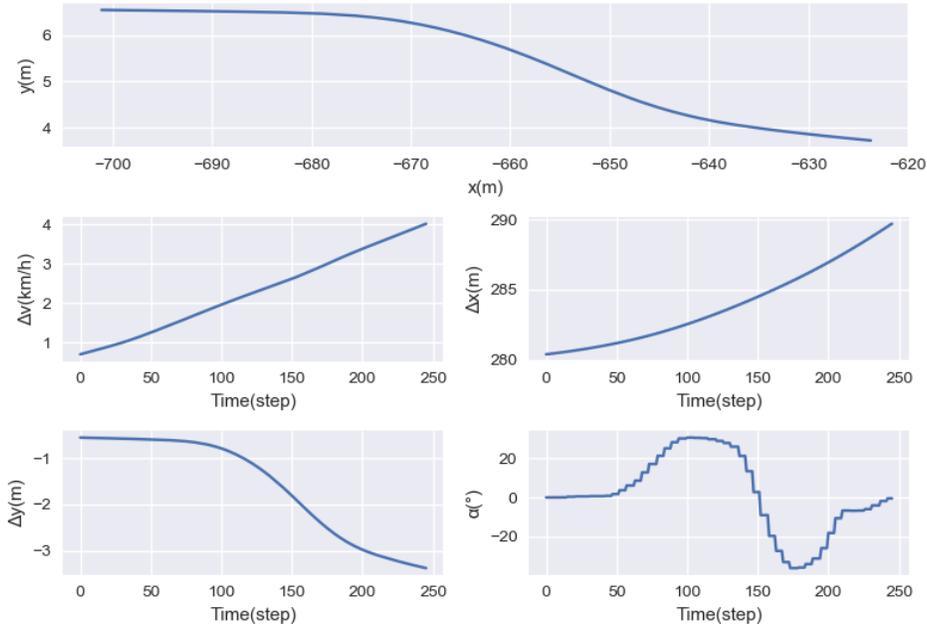

Figure 6   Example of lane-changing trajectory and data samples collected in the scenario with speed of 60 km/h.

In order to consider the time-series characteristic of lane-changing process, the input should be a historical sequence of variables. And for different samples, the input sequence should be the same length. Due to the heterogeneity of driving behaviors of different drivers and the influence of other factors, the lane-changing duration is different, that is to say the sequence length of the lane-changing process is different. Therefore, we split the trajectories into segments of the same length using the sliding windows method, where we use fixed-length input and output (different time steps and prediction horizons will be discussed in the next section). When validating the model, two-thirds of lane-changing events from each driver are used as the training dataset, and the other one-third of samples are considered as the testing dataset. Additionally, we randomly take 20% of the samples from the training dataset as the validation dataset.

## 4. Modeling Evaluation and Analysis

In order to explore the impact of different input dimensions and various variables on the prediction performance of the TCN. We first predict the lane-changing behavior and trajectory separately, and perform a sensitivity analysis using various input variables for behavior prediction. Here, we take the lane-changing data samples collected in the scenario with speed of 60 km/h as an example to introduce the learning process and results of the proposed method. Finally, the prediction performance of the joint lane-changing behavior and trajectory model is examined. All the training and testing process have been executed on a computer using the TensorFlow framework. The operative system is a Windows 10 over an Intel(R) Core (TM) i7-9700 CPU at 3.00 GHz de 16 GB of RAM and a NVIDIA GeForce RTX 2070 SUPER with 8 GB of RAM.

### 4.1 Lane-changing behavior prediction

To demonstrate the ability of proposed model to predict long-term behaviors, we need to explore the performance of the model for different prediction horizons. Before that, we first determine the optimal input length (time step) corresponding to the optimal result for each prediction horizon, where $time\ step \in \{10, 30, 50, 80, 100\}$ and $prediciton\ horizon \in \{1, 10, 30, 50, 80, 100\}$. The maximum value of prediction horizon is set 100 because there are only 224 data points in the lane-changing event with the shortest duration. The leftmost bar in each subgraph of Figure 7 shows the MSE of the TCN for different prediction horizons and time steps. We select the time steps corresponding to the smallest MSE and then we can determine the sample size accordingly in each prediction horizon. The second column of Table 1 shows the optimal time step and sample size of each prediction horizon.

The TCN is established using the best model weights saved during training and optimal time step. As the TCN belongs to neural network method, we compare prediction performance of the proposed method with two neural network methods: Convolutional Neural Network (CNN) and Recurrent Neural Network (RNN). The CNN has a similar convolution structure of the TCN. And the RNN is one commonly used method for analyzing the time series. Their optimal time step and sample size are also determined separately like the TCN, which is shown in Figure 7 and Table 1. In addition, we use the same training and testing data as the TCN. All models in different prediction horizons are trained to converge. Figure 8 shows the training process of prediction horizon = 100.

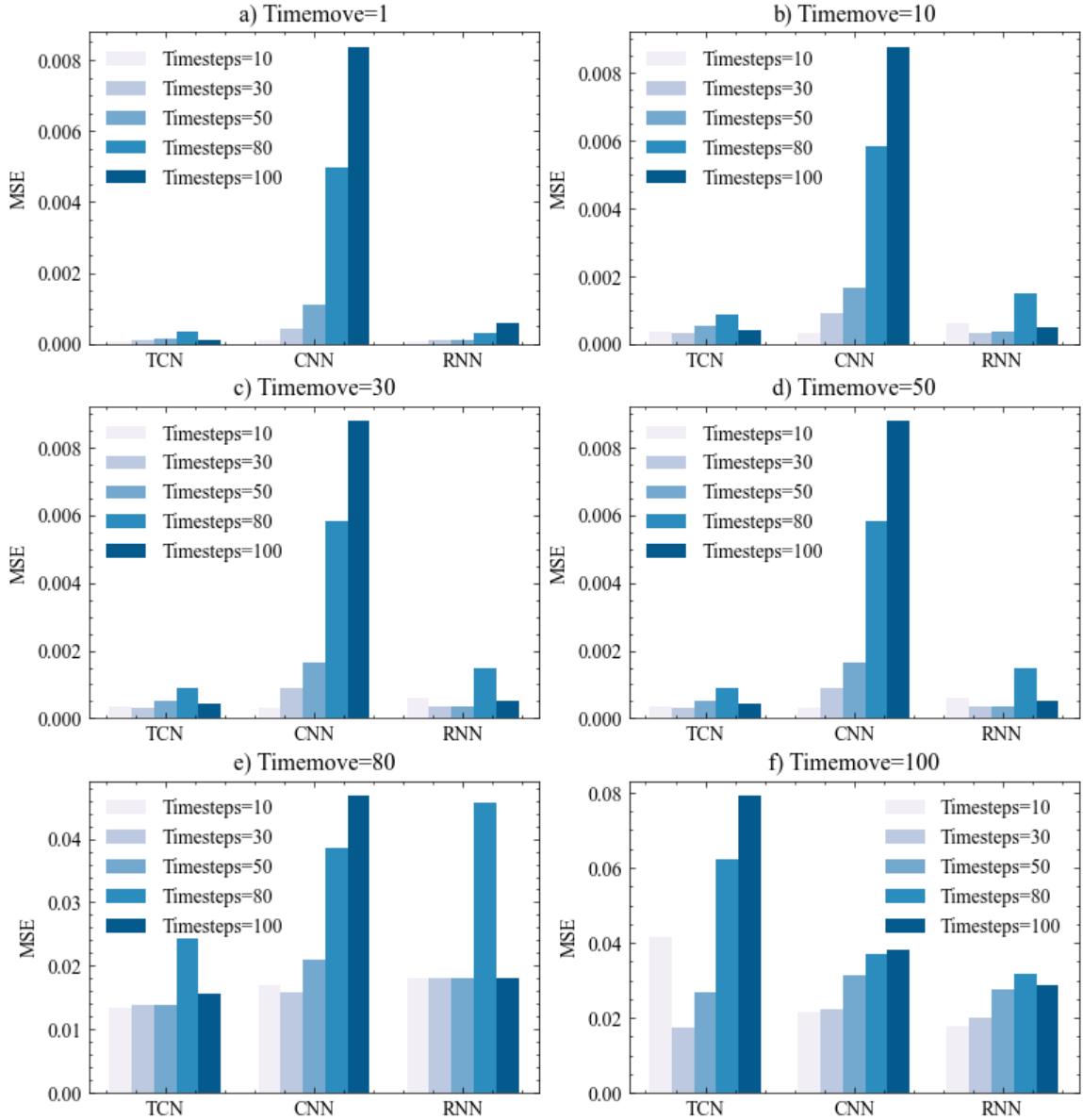

Figure 7 MSE of the TCN, RNN and CNN behavior prediction on different prediction horizons and time steps

Table 1 Sample size and optimal input time step of behavior prediction corresponding using different prediction horizons

| Prediction horizon | TCN | | RNN | | CNN | |
|---|---|---|---|---|---|---|
| | Time step | Sample size | Time step | Sample size | Time step | Sample size |
| 1 | 10 | 22574 | 10 | 22574 | 10 | 22574 |
| 10 | 30 | 21240 | 30 | 21240 | 10 | 22160 |
| 30 | 30 | 20320 | 30 | 20320 | 10 | 21240 |
| 50 | 30 | 19400 | 30 | 19400 | 10 | 20320 |
| 80 | 10 | 18940 | 30 | 18020 | 30 | 18020 |
| 100 | 30 | 17100 | 10 | 18020 | 10 | 18020 |

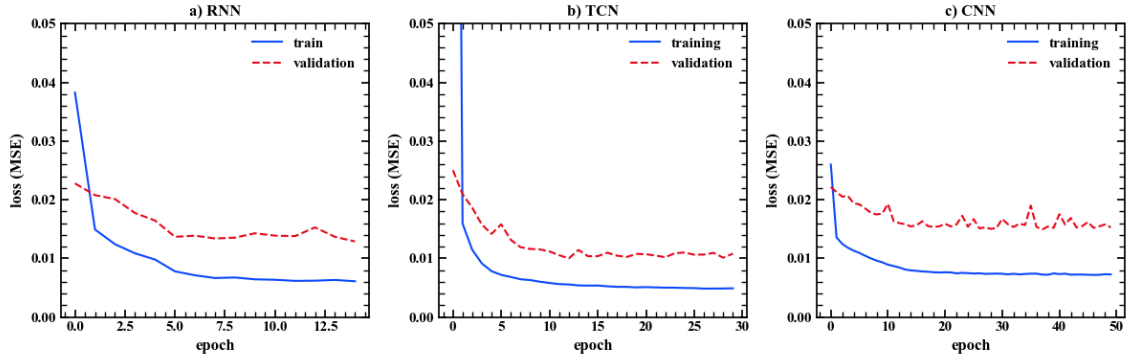

Figure 8 Learning curves of MSE loss for models (prediction horizon=100)

Figure 9 shows the prediction results of steering wheel angle $\alpha$ during a lane-changing event using the trained TCN when prediction horizon =1. In this figure, dotted line represents the actual data collected from simulator. The red line, green line and blue line represent the $\alpha$ value predicted by the TCN, RNN and CNN respectively. These three models have good prediction performance in the short term. As can be seen from the partially enlarged view on the right of Figure 9, the $\alpha$ values predicted by the TCN are closer to the actual values. Figure 10 shows the prediction results of lateral position *y* in several lane-changing segment using the trained TCN trajectory model when prediction horizon =100. As we can see, the TCN can accurately capture the sudden changes in the short-term and long-term uneven changes of steering wheel angle.

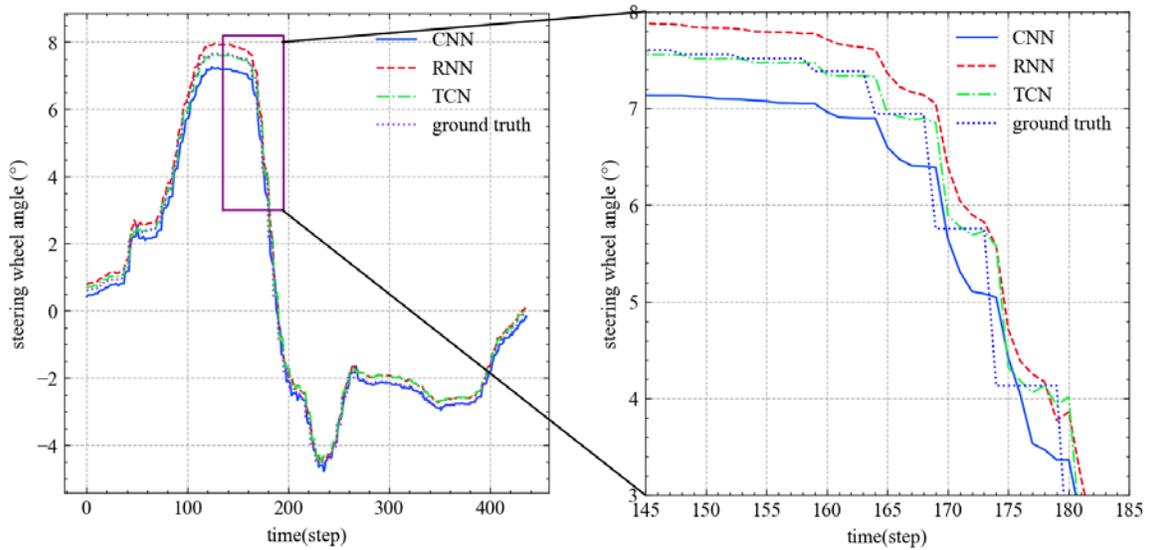

Figure 9  Prediction results of behavior for a lane-changing event (prediction horizon =1)

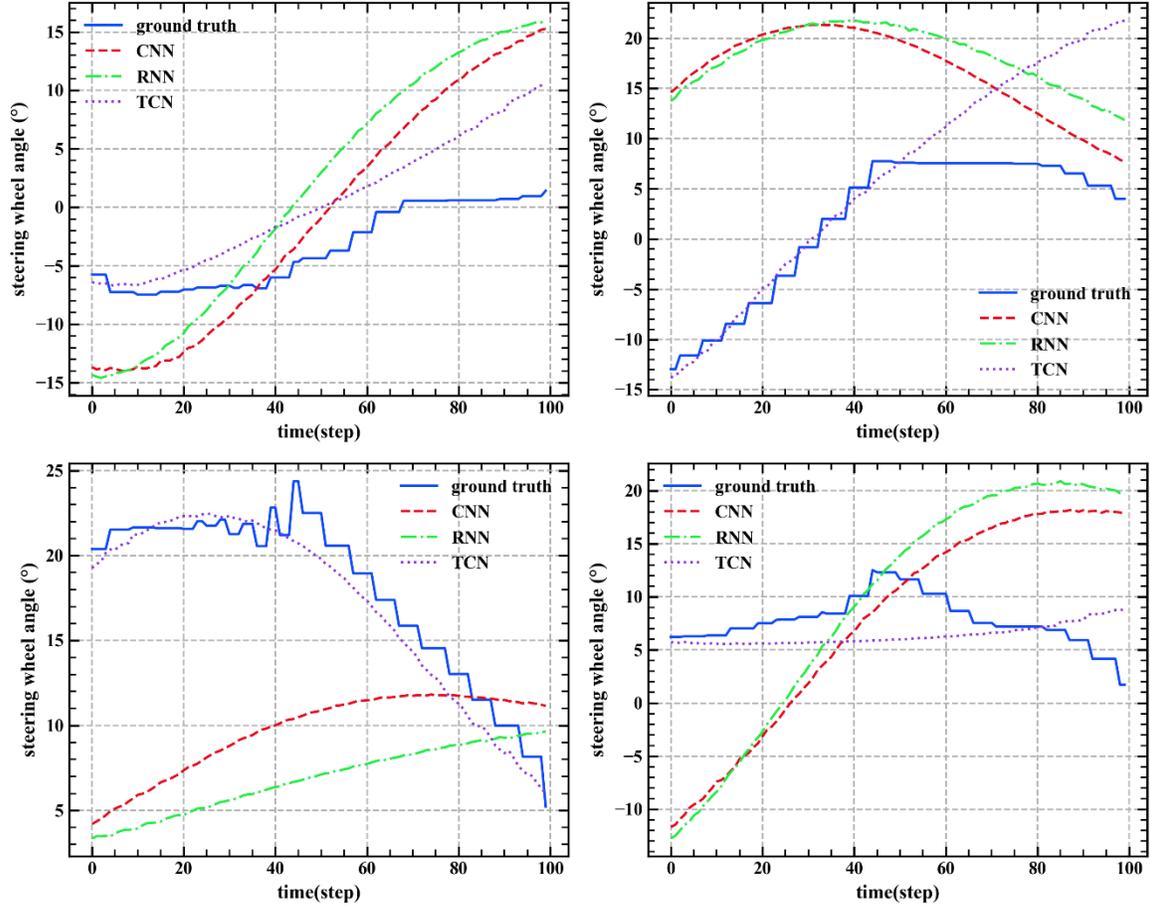

Figure 10  Prediction results of behavior for a lane-changing event (prediction horizon=100)

Furthermore, we define two indicators as the criteria for evaluating the effects of the three models on different prediction horizons: mean absolute error (MAE) and MSE. The definition of MAE can be seen in Equation 14 and the definition of MSE can be seen in Equation 6.

$$MAE = \frac{\sum_{k=1}^{N}\sum_{t=1}^{T}\left|\hat{Y}_{ti} - Y_{ti}\right|}{T \times N} \qquad (14)$$

Table 2 shows prediction accuracy of the three methods on different prediction horizons according to MAE and MSE. In the process of selecting the prediction horizon from 1 to 100, the errors of the three models are increasing. This is because as time increases, the time dependence of the current state on the historical state gradually decreases. However, the prediction errors of the TCN grow slowly and its accuracy in long-term prediction is the highest. The reason is that the structure of the dilated convolution in the TCN could build very long effective historical dependence without leakage. Followed by the TCN, the CNN has the least performance on long-term prediction because it captures the similarity between adjacent data points instead of time-series characteristics. For the RNN, it has the best performance when prediction horizon is 1, while it has worse performance on long-term performance. The simple RNN is not good at capturing long dependencies, because in the back-propagation process, the gradient from the output is difficult to propagate back and affect the weight of the previous layer. Therefore, the output of RNN is affected by the input that is closer to the current state.

Table 2 Prediction results of behavior models using different prediction horizons

| Prediction horizon | Prediction accuracy ($*10^{-4}$) | Prediction methods | | |
|---|---|---|---|---|
| | | TCN | RNN | CNN |
| 1 | MSE | 0.88 | **0.77** | 1.27 |
| | MAE | **42.16** | 54.42 | 75.42 |
| 10 | MSE | **3.05** | 3.41 | 3.16 |
| | MAE | **109.4** | 115.24 | 127.73 |
| 30 | MSE | 24.16 | 33.66 | **22.83** |
| | MAE | **304.34** | 402.22 | 309.73 |
| 50 | MSE | **57.93** | 81.61 | 58.21 |
| | MAE | **482.26** | 631.57 | 506.12 |
| 80 | MSE | **133.32** | 179.69 | 158.28 |
| | MAE | **786.23** | 969.55 | 891.75 |
| 100 | MSE | **177.16** | 198.65 | 206.06 |
| | MAE | **922.26** | 977.75 | 1039.16 |

For quick lane changes, short computational time for prediction model is necessary for the development of reliable ADAS. Table 3 displays the computational time of models (The batch sizes of model testing are the same, so the time of each step is compared). The computational time can reflect the complexity and response speed of the model to a certain extent. It can be observed that the accuracy of TCN is not only based on improving the complexity of the model. The models with convolution structure has shorter testing time while the RNN has longer computational time for larger prediction horizons, because the recurrence is not parallelizable in nature. Therefore, considering the accuracy and computational cost, the TCN is more applicable for real-time lane-changing trajectory prediction.

Table 3 Computational time of behavior models for different prediction horizons

| Prediction horizon / Computational time (us/step) | Prediction methods | | |
|---|---|---|---|
| | TCN | RNN | CNN |
| 1 | 261 | 395 | 167 |
| 10 | 283 | 762 | 208 |
| 30 | 306 | 797 | 214 |
| 50 | 332 | 807 | 255 |
| 80 | 373 | 873 | 283 |
| 100 | 265 | 455 | 122 |

In our dataset, there are also relative indicators between subject vehicle and other vehicles. However, these indicators are not contained in our study. Accordingly, we employ a sensitivity analysis to further study the influence of subject vehicle parameters on prediction results in the long term (prediction horizon =100). Table 4 provides the results of behavior prediction using the TCN by adding input variables $\Delta x$, $\Delta y$ and $\Delta v$. It can be observed firstly that including additional parameters can result in the deterioration of prediction performance. It can be also found that considering the relative distance in $y$ direction in the prediction model can significantly affect the prediction results.

Table 4 Sensitivity analysis of variables for the TCN behavior model

| **Relative indicators removed** | MSE | MAE |
|---|---|---|
| $\Delta x$ | 0.0195 | 0.1009 |
| $\Delta y$ | 0.0581 | 0.1963 |
| $\Delta v$ | 0.0207 | 0.1023 |

## 4.2 Lane-changing trajectory prediction

Different from predicting lane-changing behavior, trajectory prediction is a two-dimensional sequence modeling problem. Thus, the sample sizes of different models in different prediction horizon are the same, see Table 5.

Table 5 Sample size of trajectory models corresponding to different prediction horizons

| **Prediction horizon** | **Sample size** |
|---|---|
| 1 | 22988 |
| 10 | 22160 |
| 30 | 20320 |
| 50 | 18480 |
| 80 | 15720 |
| 100 | 13880 |

Similarly, the prediction performance of the RNN and CNN are also compared with the proposed method. Table 6 shows the prediction accuracy of the three methods for the trajectory data with different prediction horizons. Similar to the prediction results of the lane-changing behavior data, the advantages of TCN become more obvious as the prediction horizon increases.

Table 6 Prediction results of trajectory data with different prediction horizons

| Prediction horizon | Prediction accuracy ($*10^{-4}$) | Prediction methods | | |
|---|---|---|---|---|
| | | TCN | RNN | CNN |
| 1 | MSE | 0.88 | 24.57 | **0.01** |
| | MAE | 42.25 | 416.56 | **6.89** |
| 10 | MSE | 2.09 | 27.43 | **1.47** |
| | MAE | 90.94 | 445.03 | **68.88** |
| 30 | MSE | **11.77** | 39.13 | 12.93 |
| | MAE | **225.24** | 515.2 | 205.7 |
| 50 | MSE | **34.38** | 65.03 | 40.09 |
| | MAE | **343.56** | 619.64 | 348.54 |
| 80 | MSE | **55.31** | 123.71 | 92.63 |
| | MAE | **417.99** | 811.45 | 536.93 |
| 100 | MSE | **97.92** | 137.77 | 127.61 |
| | MAE | **622.33** | 856.67 | 641.13 |

**4.3 Lane-changing behavior and trajectory prediction**

Accurate lane-changing prediction is beneficial for drivers to make decisions in advance. In this section, a joint prediction model which simultaneously predicts long-term lane-changing behavior and trajectory is developed (prediction horizon=100 time steps). In order to demonstrate the advantages of the proposed method, the RNN and CNN are also used for comparison and the performance of models in three different scenarios are shown. Table 7 shows prediction accuracy of methods under different speed conditions according to the two error indicators. As can be seen from the table, due to the dilated convolution structure and residual connections of the TCN, the long-term prediction accuracy of the TCN is higher than the RNN and CNN in scenarios with different speed. Taking MSE as an example, the prediction accuracy of the TCN increases by 35%-65% compared to the RNN and 35%-40% compared to the CNN.

Furthermore, the prediction results of each variable of the three models are shown in Table 8. The CNN has the best prediction performance for longitudinal position, while the TCN has the best prediction performance for lateral position and steering wheel angle. The advantage of CNN for longitudinal position prediction is that the ordinary convolution structure of CNN can accurately analyze the simple linear trend of longitudinal position. As discussed in the previous studies, the nonlinear and stochastic lateral position and steering wheel angle are more important in the lane-changing process. Compared with the CNN and RNN, the TCN can capture long-term time-series characteristics of the lateral position and steering wheel angle. Overall, the TCN can outperform the RNN and CNN in multi-dimensional sequence prediction.

Table 7 Prediction results of models in scenarios with different speed

| Driving speed | Prediction accuracy (*10⁻⁴) | Prediction methods | | |
|---|---|---|---|---|
| | | TCN | RNN | CNN |
| 60km/h | MSE | **186.45** | 284.21 | 287.74 |
| | MAE | **931.93** | 1287.41 | 1160.78 |
| 80km/h | MSE | **56.21** | 158.62 | 95.35 |
| | MAE | **536.98** | 943.75 | 647.05 |
| 100km/h | MSE | **53.60** | 131.42 | 89.59 |
| | MAE | **503.95** | 840.72 | 615.77 |

Table 8 MSE (*10⁻⁴) for each variable in scenarios with different speed

| Driving speed | Variables | Prediction methods | | |
|---|---|---|---|---|
| | | TCN | RNN | CNN |
| 60km/h | Longitudinal position | 2.20 | 24.36 | **2.05** |
| | Lateral position | **203.26** | 276.01 | 309.93 |
| | Steering wheel angle | **353.87** | 552.25 | 557.54 |
| 80km/h | Longitudinal position | 14.12 | 60.32 | **1.96** |
| | Lateral position | **55.16** | 198.97 | 136.90 |
| | Steering wheel angle | **99.35** | 216.57 | 147.18 |
| 100km/h | Longitudinal position | 5.69 | 50.73 | **0.21** |
| | Lateral position | **86.70** | 152.50 | 129.18 |
| | Steering wheel angle | **68.41** | 191.02 | 139.40 |

## 5. Conclusion

Accurate lane-changing prediction is an important component for developing a reliable and efficient ADAS. Lane changes are complicated and stochastic due to human factors, so it is difficult to adequately capture the dynamic of lane changes using kinematic models and traditional machine learning method such as SVM. For freeway lane changes, the drivers can perceive the potential dangerous vehicle movements using the long-term prediction results from the TCN.

In this study, a temporal convolution network is introduced to predict lane-changing behavior and trajectory in the long term. We define the steering wheel angle as the indicator of the lane-changing behavior and the two-dimensional position as the indicator of the lane-changing trajectory. Considering the characteristics of lane changes and the advantages of deep learning approach, we proposed the TCN to predict lane-changing behavior and trajectory in the long term. The CNN and RNN based model are compared to demonstrate the superior learning ability of the TCN. Besides, we demonstrate the prediction performance of TCN from three aspects: different input variables, different input dimensions and different driving speeds. The results demonstrate that the proposed model is able to accurately and rapidly predict the lane-changing process of a vehicle, and it outperforms the CNN and RNN method, especially for the long-term prediction. The TCN considers the time-series characteristics of lane changes, using its dilated convolution structure to expand the receptive field to capture longer-term historical information and enhance memory ability. In summary, the proposed model can provide accurate long-term prediction with less computational time, which is useful for predicting the freeway lane change events.

For future research, this study can be expanded from the following aspects: First, other traffic scenarios (e.g., merging areas and weaving areas) should also be examined. Second, the lane change data used in this study are the high-frequency data collected by the driving simulator. The real-world traffic data can be collected to further verify the performance of the proposed model. Third, in the sensitivity analysis, we consider the relative indicators between the subject vehicle and lead vehicle. In the future, the indicators about vehicles in the target lane should be taken into account in the model.

**Acknowledgments**

This research was funded by the National Natural Science Foundation of China (Grant No. 71971160), the Shanghai Science and Technology Committee (Grant No. 19210745700) and the Fundamental Research Funds for the Central Universities (Grant No. 22120200035).